\documentclass[runningheads]{llncs}
\usepackage[T1]{fontenc}
\usepackage{hyperref}
\usepackage{graphicx}
\usepackage{amsmath}
\usepackage{amssymb}
\usepackage{wrapfig}
\usepackage{multirow}
\usepackage{booktabs}
\usepackage{xcolor}
\usepackage{here}
\usepackage{rotating}
\usepackage{longtable}

\usepackage{titlesec}
\titlespacing{\section}{0pt}{0.6\parskip}{0.4\parskip}
\titlespacing{\subsection}{0pt}{0.5\parskip}{0.3\parskip}

\begin{document}

\title{Improved Models for Media Bias Detection and Subcategorization}

\titlerunning{Improved Media Bias Model}

\author{Tim Menzner\inst{1}${}^\ast$ \and Jochen L.~Leidner\inst{1,2}\orcidID{0000-0002-1219-4696}} %
\authorrunning{T.~Menzner and J.~L.~Leidner}

\institute{Information Access Research Group, Coburg University of Applied Sciences, Friedrich-Streib-Straße 2, 96459 Coburg, Germany \and University of Sheffield, Department of Computer Science, Regents Court, 211 Portobello, Sheffield S1 4DP, United Kingdom\\ ${}^\ast$ Correspondence: \email{tim.menzner@hs-coburg.de} / \email{leidner@acm.org}}

\maketitle

\begin{abstract}
  We present improved models for the granular detection and sub-classification news media bias in English news articles.
  We compare the performance of zero-shot versus fine-tuned large pre-trained neural transformer language models,
  explore how the level of detail
  of the classes affects performance on a novel
  taxonomy of 27 news bias-types, and demonstrate how using synthetically generated example data can be used to improve quality.
  
  %\and pre-trained neural transformers \and large language models \and information retrieval \and 
\end{abstract}

%%%%%%%%%%%%%%%%%%%%%%%%%%%%%%%%%%%%%%%%%%%%%%%%%%%%%%%%%%%%%%%%%%%%%
% Introduction
%%%%%%%%%%%%%%%%%%%%%%%%%%%%%%%%%%%%%%%%%%%%%%%%%%%%%%%%%%%%%%%%%%%%%

\section{Introduction} \label{sec:intro}

\parindent0em
Unbiased, trusted news reporting is crucial for
sustaining democratic political
systems, yet news media are
exposed to manipulation leading
to news bias from the outside
(propaganda) as well as the
inside (agenda of the owners of
a news medium).
This paper is part of a line of
work researching questions of
to what degree and how computers
can automatically detect instances
of news media bias and categorize
them into sub-classes. Any model
of bias capable of this ought to
be enormously valuable, since
its use as a predictive device
can assist humans by alerting them
to instances of bias in reporting.

Media bias can be described as the tendency to, consciously or unconsciously,
report a news story in a way that supports a pre-existing narrative instead of
providing unprejudiced coverage of an issue.
In contrast, ``[p]ropaganda is neutrally defined as a systematic form of purposeful persuasion that attempts to influence the emotions, attitudes, opinions, and actions of specified target audiences for ideological, political or commercial purposes through the controlled transmission of one-sided messages (which may or may not be factual) via mass and direct media channels.'' \cite[pp. 232-–233]{Nelson:1996}, so unlike bias, propaganda is always intentional.
We are concerned particularly with media bias in English-language, online, written news here; while other
languages are of equal importance
(and, in practice, often neglected), English is commonly
used as the benchmark language
to compare NLP models for various tasks on, and resources such as annotated corpora are more readily available.

A range of authors -- perhaps beginning with \cite{Lin-etal:2006:CoNLL} -- have addressed
the news bias modeling question before (Section \ref{sec:related-work} below addresses related work), but since the availability of pre-trained neural transformer models (often just called ``Large Language Models'', LLMs), the
quality of automated predictive
models for NLP tasks has increased in general \cite{Devlin-etal:2019:NAACL,Vaswani-etal:2017:NeurIPS}, and this has
in turn led to better news bias
models \cite{Hamborg:2023,Menzner2024}. 
We will address a number of research questions, and aiming to
answer them will help us improve
our understanding of news bias
modeling with neural transformers,
which in turn will lead to better models.

\textbf{Research Questions and Contributions.}
We address the following research questions: \\
    %\sloppypar{}
    \textbf{RQ-1.} \textit{How does the level of detail of the categories impact the ability to identify instances of news bias?}
    News bias detection is already
    a difficult task for humans
    and machines, so having fine-grained sub-classes may be beyond the state of the art. On the other hand, more granular classes may help the
    classifier distinguish better between different cases. \\
    %\sloppypar{}
    \textbf{RQ-2.} \textit{Can we find conditions on which LLM hallucination depends on?} %weiß nicht ob die frage nicht etwas zu weit ist, im grunde zeigen wir ja nur dass finetuning die halluzinationen nicht reduziert. Ob das als Antwort für so eine weite Frage ausreicht, weiß ich nicht -tim
    % Stimmt alles. Man kann auch
    % Teilantworten geben zu einer
    % Frage wenn man 4 RQ hat.
    % können wir auch später ersetzen durch eine andere
    % Frage bzw. umsortieren
    %Ok
    Large language model hallucination is one of the key problems that prevent
    deployment in sensitive applications; any insights
    regarding their reduction is
    valuable. \\
    %\sloppypar{}
    \textbf{RQ-3.} \textit{How do the largest zero shot models compare to fine-tuned models?}
    Traditionally, supervised models have been superior to
    unsupervised models; large, pre-trained transformers have
    tipped the scale towards training with raw text. Are carefully fine-tuned models superior to larger, non-fine tuned models?\\
    %\sloppypar{}    
    \textbf{RQ-4.} \textit{How does synthetic data augmentation help for improving the task?} Language models can be used to generate training data for additional training (fine-tuning). Can (and if so, how) can such data augmentation help to improve overall performance on the task of news media bias detection and sub-classification?
    
Our contributions are (i) answers to these research questions supported by  detailed experimental evaluations on multiple datasets and (ii) a novel, very granular taxonomy of 27 news bias-types, and (iii) a set of new models for the improved  detection and sub-categorization
of news bias in English-language media.

%%%%%%%%%%%%%%%%%%%%%%%%%%%%%%%%%%%%%%%%%%%%%%%%%%%%%%%%%%%%%%%%%%%%%
% Related Work
%%%%%%%%%%%%%%%%%%%%%%%%%%%%%%%%%%%%%%%%%%%%%%%%%%%%%%%%%%%%%%%%%%%%%

\section{Related Work} \label{sec:related-work}

\subsection{News Bias: The Phenomenon}

% Media bias \cite{Lee-Solomon:1990,Groseclose-Milyo:2005:QuartJEcon,Sloan-Mackay:2007,Groeling:2016:AnnuRevPolitSci} has a
% long tradition of being investigated, and often,
% readers are aware of the political leanings of 
% a newspaper, online or in paper form. However, the
% increasing use of the Internet as a target for
% information warfare and propaganda has led to an
% increase of various types of bias, fake news and
% other undesirable phenomena.
% Evidence also suggests \cite{Vosoughi-Roy-Aral:2018:Sci} that fake news spreads
% faster than balanced news.
% DellaVigna et al. \cite{DellaVigna:2007} focused on the effect of news bias and voting patterns, suggesting a significant increase in republican votes in towns where Fox News entered the cable market in 2000. 
% Groeling \cite{Groeling:2016:AnnuRevPolitSci} presents a survey of the literature covering partisan bias.
% However, studying political bias in U.S. news reporting,
% Budak et al. \cite{Budak-etal:2016:PublOpinQuart} found that ``news outlets are considerably more similar than generally believed. Specifically, with the exception of political scandals, major news organizations present topics in a largely nonpartisan manner, casting neither Democrats nor Republicans in a particularly favorable or unfavorable light.'' 

A stage model to explain the arise of media bias during the news production process was proposed by Hamborg et al. \cite{Hamborg2018}. The authors describe how bias can be introduced by several factors like the political views of news producers or the demand of a target audience during different steps like the information gathering and the writing.
Martin et al. \cite{Martin2017} investigated the impact of media bias on voting behavior and consumer preferences for news aligned with their own ideology, finding that additional weekly viewership of a channel can slightly increase the probability of intending to vote for the political side associated with its bias.

\subsection{Automatic Identification of Bias in News}

One of the first approaches
of automatic news bias detection
was described by Lin \textit{et al.} \cite{Lin-etal:2006:CoNLL}.
Gentzkow et al. \cite{Gentzkow2010} compared phrases and words predominantly used by members of the US Congress of one political party with the language used in news media coverage to identify political bias.
A combination of traditional NLP techniques and OpenAI Inc.'s GPT-4.0 was used by \cite{Benson2024} to analyse topics discussed in cable news media and their respective stance towards it in order to provide a general assessment of its bias. The authors found that such a stance based approach was superior to a one solely focused on sentiment.
Mancini et al. focused on multi-modal fallacy classification in political debates, which could be viewed as a specific sub-type of news bias \cite{mancini-etal-2024-multimodal}.
Datasets for news bias on a sentence level as well as evaluation of detection approaches were provided by \cite{Spinde2021f} and \cite{fan-etal-2019-plain}. Recently, Nakov's research group
\cite{martino2019finegrained,baly-etal-2020-detect} published a BERT based system to detect 18 different propaganda techniques in news articles, along with the respective  annotated data set. While news bias is a broader phenomena than propaganda (as it also includes unintentional subjective reporting), both issues are related, as visible in the overlap of the identified propaganda techniques in this work and the bias-types discussed in ours.

\subsection{Categorization of Different Types of Bias} \label{Related Work Category}

Rodrigo-Ginés \textit{et al.} \cite{RodrigoGines-CarrilloDeAlbornoz-Plaza:2024:ExpSysAppl} 
 identified 17 different types of media bias, depending on context and intention, based on reviewing the existing literature. A more coarse-grained category inventory was given by \cite{Wessel-etal:2023:SIGIR}, where 9 types of bias were used for the construction of their dataset. An overview with 16 identifies types of media bias and examples was presented by \cite{Mastrine:2019:online}. Most recently, \cite{Silfwer:2020:online} presented a list of common logical fallacies and cognitive biases, which arguably also play a role in reporting.
Very related to bias is propaganda; {Da San Martino} \textit{et al.} present
a fine-grained taxonomy and classification model for detecting propaganda types that complements this work on bias \cite{da-san-martino-etal-2019-fine}.

\section{Preliminaries}

\subsection{Bias Categorization} \label{bias_types}
Media bias can be categorized into two types: visible bias within an article or sentence, and "Meta-Bias," which stems from the broader context. Examples of Meta-Bias include a news outlet's tendency to prioritize certain stories (known as "gatekeeping bias"), their placement, and their allotted space \cite{Hamborg2018}. Detecting this kind of bias is hard as it requires a wider knowledge of context and publication history of an outlet.
This paper concentrates on detecting sentence-level bias, which occurs within individual sentences. Building upon related work and interacting and experimenting with sentence categorization using GPT, along with our own observations, we identified 27 types of bias as follows:

\begin{multicols}{2}
\begin{small}
\begin{description}
    \item[Ad Hominem Bias] targeting the human (the character, motives, or other attributes of the one making the argument) rather than the argument itself
    \item[Ambiguous Attribution Bias] a position is broadly attributed to a wide, unspecified group such as "experts", "economists", or "politicians", rather than to identified individuals/sources
    \item[Anecdotal Evidence Bias] relying on individual stories or examples rather than considering broader, more representative evidence
    \item[Causal Misunderstanding Bias] a cause-and-effect relationship between two variables is misunderstood or assumed without sufficient evidence or considering other factors
    \item[Cherry Picking Bias] giving undue prominence to aspects of a news story that endorses a certain viewpoint while omitting information that would contest it
    \item[Circular Reasoning Bias] the conclusion of a statement or argument is used as its own justification
    \item[Commercial Bias] emphasizing or promoting certain products, services, or narratives due to underlying commercial interest
    \item[Discriminatory Bias] promoting stereotypes, generalized or prejudiced statements and unequal representation, reinforcing discrimination against certain individuals or groups, often based on ethnicity, culture, nationality, social background, gender, sexual orientation, or religious beliefs
    \item[Emotional Sensationalism Bias] using hyperbolic or provocative language designed to evoke (strong) emotions, usually at the expense of accuracy or context while often focusing predominantly on negative events, aspects, or interpretations
    \item[External Validation Bias] deeming something valid or true simply because it is supported by an authority figure or because it aligns with the beliefs or actions of a large group of people
    \item[False Balance Bias] presenting opposing viewpoints as equally credible or significant, despite a clear consensus or evidence favoring one side
    \item[False Dichotomy Bias] presenting a complex issue as leaving only two opposing decision alternatives when there might be further possible solutions/positions/outcomes
    \item[Faulty Analogy Bias] drawing comparisons between two things that may share superficial similarities but are fundamentally different
    \item[Generalization Bias] extrapolating characteristics of a specific subset to a larger group, or conversely, attributing broad characteristics of a group to each of its individual members
    \item[Insinuative Questioning Bias] posing suggestive questions that contain implicit assumptions or lead the audience towards a pre-conceived notion, often used to promulgate subjective beliefs or doubts under the pretense of neutral inquiry
    \item[Intergroup Bias] dividing people into two groups with one group (often an in-group to which the writer or publication belongs or identifies with) and portraying one as positive, while a second group, the out-group, is attributed negative characteristics and seen as adversarial
    \item[Mud Praise Bias] using personal attacks, rumors, or unfounded allegations to damage the reputation of an individual or a group, or the opposite tendency to excessively praise or idealize them without regard for objective assessment
    \item[Opinionated Bias] including subjective material, but portrayed as objective reporting; obscuring the line between fact and personal perspective
    \item[Political Bias] inclination towards a specific political party, ideology, or candidate, typically resulting in favoritism towards one side while disregarding or disparaging opposing viewpoints
    \item[Projection Bias] attributing thoughts, feelings, motives, or intentions to others (be it individuals, groups, or entities) without sufficient evidence or direct statements to back such claims
    \item[Shifting Benchmark Bias] changing an argument, e.g., in response to criticism, by excluding counterexamples or adjusting the criteria to maintain a certain outcome
    \item[Source Selection Bias] citing sources that likely are themselves biased with respect to the topic
    \item[Speculation Bias] speculating based on conjecture about situations or outcomes rather than relying on concrete facts and definitive evidence
    \item[Straw Man Bias] misrepresenting / distorting an argument so as to make it easier to attack, e.g., by oversimplifying or exaggerating
    \item[Unsubstantiated Claims Bias] presenting statements or assertions as factual without providing adequate evidence or references
    \item[Whataboutism Bias] deflecting or responding to an accusation or problem by making a counter-accusation or raising a different issue, not addressing the original argument
    \item[Word Choice Bias] words with inherent positive or negative connotations, euphemisms, dysphemisms, or strong adjectives are chosen, influencing perceptions and implying judgment about a subject
\end{description}
\end{small}
\end{multicols}

\vspace{-0.55cm}
Different bias types often intertwine; for instance, Political Bias may coincide with Word Choice Bias and Opinionated Bias. They can be categorized together or separately based on desired precision. For example, Casual Misunderstanding Bias and False Dichotomy Bias could be seen as subsets of "Logical Fallacy Bias". Similarly, Casual Misunderstanding Bias might further branch into various types, such as confusing causation with correlation or falling into the Prevention Paradox (Rose, 1981). While all bias-types are detectable from single sentences alone, some like Cherry Picking Bias would benefit from further context. While it be can be obvious from a sentence alone, like solely emphasizing the positives of a highway project near a nature reserve, without addressing any environmental concerns, more intricate cases, such as selectively mentioning a protest without acknowledging a larger counter protest, necessitate specific event knowledge for identification.

Contrary to the assessment presented in \cite{RodrigoGines-CarrilloDeAlbornoz-Plaza:2024:ExpSysAppl}, we further assume that media bias does not have to be intentional but can also be introduced subconsciously (in contrast, propaganda is never subconscious). 

\subsection{Bias Strengths}
Bias is not necessarily a binary classification but could be described as a spectrum. A sentence may not be just biased or unbiased but be more biased or less biased. When a bias is subtle, it might be not so obvious too detect but could still influence a readers opinion. We therefore suggest rating the bias strength of a sentence on a scale from 0.0 (no bias at all) to 1.0 (very extreme bias). 

\subsection{Prompting}
\label{prompting}
The media bias definition from Section \ref{sec:intro} was used, along with the bias-types to develop a LLM prompt for bias detection. In order to improve results and provide a benchmark for the bias strengths, each bias-type was enhanced by providing two example sentences of different bias strengths. During testing, the substitution of the decimal numbers with descriptive words (such as negligible, noticeable or significant) was also tried, but as it did not appear to affect the model decision, the decision was made to stick with numbers. An example of a type definition with example sentences is given in Example \ref{ex:1}:

\begin{example} \label{ex:1} \hrule Examples for a bias-type definition used for prompting\\
Insinuative Questioning Bias: This is the practice of posing suggestive questions that contain implicit assumptions or lead the audience towards a preconceived notion, often used to promulgate subjective beliefs or doubts under the pretense of neutral inquiry. \\
Examples: \\
Moderate Bias Strength (0.6): "Does their community improvement plan also serve a political purpose for his campaign?" \\
High Bias Strength (0.9): "Isn't the so-called community improvement plan just a ploy by them to fool voters before the election?"
\hrule
\end{example}

%%remember to include conversely in definition of generalization bias

When formulating the example sentences, it was tried to avoid referencing specific real world issues or topics when possible, in order to not introduce any form of bias regarding those. We further applied known best practices, like asking the model to assume a role as an expert in media bias and describing the task (identifying, categorizing and rating biased sentences, providing an overall assessment and returning the results in JSON format)  step by step \cite{openaiOpenAIPlatform}.
%%%%%%%%%%%%%%%%%%%%%%%%%%%%%%%%%%%%%%%%%%%%%%%%%%%%%%%%%%%%%%%%%%%%%
% Data
%%%%%%%%%%%%%%%%%%%%%%%%%%%%%%%%%%%%%%%%%%%%%%%%%%%%%%%%%%%%%%%%%%%%%
\section{Data} \label{sec:data}
Two publicly available datasets where used for this paper. BABE \cite{Spinde2021f} is based on MBIC \cite{Spinde2021MBIC} and includes an additional 2,000 sentences, resulting in a total number of 3700 from 14 different US news outlets. Unlike MBIC, where the labelling was done by pure crowd sourcing, all annotators for BABE had to meet certain criteria to prove a certain level of expertise. After removing the sentences where annotators could not reach an agreement, 1863 sentences labeled as non biased and 1810 labeled as bias remained in the dataset. 
The second dataset was BASIL \cite{fan-etal-2019-plain}, which contains 300 news articles about 100 different stories (from the New York Times, the Huffington
Post and FOX News), with annotations for each sentence featured in each article. Of the total 7,984 sentences in the dataset, 1727 were labeled as biased and 6257 as unbiased. 
Next to this publicly available datasets, we also used synthetic data, generated with GPT-4.0, for fine-tuning. 
To our knowledge, there are no existing News Bias Datasets including a sufficient categorisation of bias-types and bias strengths.

%%%%%%%%%%%%%%%%%%%%%%%%%%%%%%%%%%%%%%%%%%%%%%%%%%%%%%%%%%%%%%%%%%%%%
% Method
%%%%%%%%%%%%%%%%%%%%%%%%%%%%%%%%%%%%%%%%%%%%%%%%%%%%%%%%%%%%%%%%%%%%%

\section{Methods} \label{sec:method}

\subsection{Finetuning} \label{finetuning}
We fine-tuned four different models based on gpt-3.5-turbo-1106, Two with a subset of BABE and BASIL, one with synthetic data (SYNT), and one with a combination of all three (MEGA)
For BABE, BASIL and SYNT, 100 example articles were constructed from the respective data, with a randomly chosen length between 10 and 30 sentences. 
As BABE provides individual, disconnected sentences rather than complete articles, the sentences used for the fine-tune articles were randomly picked and joined together.
For BASIL, which provides full articles, the fine-tune articles were based on snippets from the articles featured in the dataset.
The synthetic articles were generated by GPT-4.0, with a random ratio of biased to unbiased sentences. The biased sentences were generated based on the type definitions and examples presented in Section \ref{bias_types}, a random distribution of desired bias strengths and an even distribution of bias-types across all articles. The unbiased sentences were generated with the instruction to be of the same topic as the biased ones. 
Based on these articles, a JSON resembling the desired output was constructed. As BABE and BASIL do not include information about bias strength or bias-type, while those were needed for our fine-tuning format, the contents of these fields were generated by GPT-4.0. 
The articles, the desired output and the later to be used prompt were combined as user message, assistant message and system prompt to end up with three fine-tuning ready datasets. A fourth one (MEGA) was constructed by appending all three files to each other.

\subsection{Experiments}
We evaluated the models on BABE and BASIL. All sentences used in the fine-tuning process were removed from both datasets beforehand. As we also compared with a previous model \cite{Menzner2024} fine-tuned on 50 example articles with 10 sentences each constructed from the MBIC dataset, which is a subset of BABE, we also removed the sentences used here. At this point, it should be noted that, as this model was not fine-tuned with our prompt including the fine-grained bias-type definitions but with a prompt making use of coarser ones \cite{Wessel-etal:2023:SIGIR}, we stuck to this prompt when evaluating this model. Because we could not rule out that single sentences from the synthetic fine-tune data were too close to sentences from one of the datasets, or that single sentences may have been modified by GPT-4-0 during the construction of the fine-tuning datasets, all removing was done via partial fuzzy string matching. In more detail, we used \cite{FuzzyWuzzy}, which compares strings based on the Levenshtein distance, to check the partial ratio of each sentence from the dataset which each sentence included in any fine-tuning data, and removed it, when a certain threshold (80) was exceeded. After this, BABE still included 1694 sentences, with an almost even split into 841 biased and 853 unbiased ones. BASIL, on the other hand, still included 4236 sentences, with a strong over representation of 3375 unbiased sentences compared to only 861 biased ones.
Based on the datasets, articles with a randomly chosen length between 10 and 30 sentences were constructed as described in Section  \ref{finetuning}. However, for BASIL, it was not always possible to reach at least 10 sentences, as some of the shorter ones had been left with less than 10 sentences all together after removing the sentences used for fine-tuning. This was true in 33 cases, with an average length of 5.97 sentences for these articles. However, as the total number of affected sentences was rather low (only 4.6\% of all evaluated sentences ended up being in a article with less than 10 sentences) and the effect of the article length on performance appears to be negligible (see Section \ref{sec:eval}), they were kept in the evaluation dataset.
These articles were then passed to the model, together with the system prompt explained in Section \ref{prompting}, using a temperature of 0.15 to maintain rigidity while also allowing for some "creativity" in answers. The sentences marked as biased by the model were then again compared to the sentences marked as biased in the datasets with partial fuzzy string matching to calculate the number of true positives, false positives, false negatives and true negatives. The partial string matching was especially required as, in order to ensure a "realistic" scenario, where the model would scan an actual article from a newspaper, the articles were passed as plain, connected text. In practice, this could result in the model picking a different sentence separation for its analysis than the sentence separation in the annotation, e.g. (not) including a introductory statement like "he told reports" before a quote. Fixed string matching is not suitable to capture these instances, partial string matching (we used a threshold of 80, meaning a sentence marked as biased by the model needed to have a ratio greater than that with anyone of the sentences marked as biased in the dataset for this article) on the other hand can do this. However, it still can not be guaranteed that each of these instances could actually be captured, which could potentially influence results.  
Beside the fine-tuned models, GPT-3.5 and GPT-4.0 were also evaluated with prompting only.
Finally, the quality of the type and strength assignment was evaluated using a manually enhanced sample from BABE and BASIL.
As previous research \cite{Menzner2024} suggested a significant worse performance of currently available Open-Source LLMs compared to the commercial ones provided by OpenAI, they were not further evaluated for these experiments.
%%%%%%%%%%%%%%%%%%%%%%%%%%%%%%%%%%%%%%%%%%%%%%%%%%%%%%%%%%%%%%%%%%%%%
% Evaluation
%%%%%%%%%%%%%%%%%%%%%%%%%%%%%%%%%%%%%%%%%%%%%%%%%%%%%%%%%%%%%%%%%%%%%

\section{Evaluation} \label{sec:eval}

%In this section, we report the results of our experimental findings.
Table \ref{table:classification_results_BABE} shows the evaluation results on the BABE dataset.
%for  GPT-3.5 respectively fine-tuned on BABE, BASIL, Synthetic Data and a combination of all three (MEGA), versus a previous model fine-tuned on MBIC and two non-fine-tuned GPT-3.5 and GPT-4.0 with prompt only. 
Overall, the fine-tuned models all outperformed the non fine-tuned GPT-3.5, with the models using a more fine-grained bias-type categorization in turn outperforming the one using a coarser one (FT MBIC). Among the fine-tuned models, FT BABE had the highest F1-score (76\%), accuracy (75\%) and precision (73\%), while FT SYNT had the highest recall (89\%). The model fine-tuned on the combined dataset also scored somewhere in the middle between the individual fine-tunes for each metric. The largest ultimate precision was achieved by GPT-4.0 (85\%), leading FT BABE by 12\%. However, as it trailed FT BABE by 16\% regarding F1-Score, by 25\% on Recall and it uses more energy, memory and comes with higher costs, FT BABE may be considered the better model overall, depending on priorities.
\begin{table}
\vspace{-10pt}
\caption{Evaluation Results on the BABE dataset for GPT-3.5-turbo-1106 fine-tuned on BABE, BASIL, Synthetic Data and a combination of all three (MEGA), a previous GPT-3.5 fine-tuned on MBIC, GPT-3.5-turbo-1106 with prompt only and GPT-4-turbo-0125. Best results are highlighted in bold.}
\label{table:classification_results_BABE}
\begin{tabular}{lrrrrrrrr}
\toprule
\textbf{Model} & \textbf{TP} & \textbf{FP} & \textbf{FN} & \textbf{TN} & \textbf{F1-Score} & \textbf{Recall} & \textbf{Precision} & \textbf{Accuracy} \\
\midrule

GPT-3.5 (FT BABE) & 576 & 214 & 154 & 524 & \textbf{0.758} & 0.790 & 0.729 & \textbf{0.749} \\
GPT-3.5 (FT BASIL) & 443 & 212 & 287 & 526 & 0.640 & 0.606 & 0.677 & 0.660\\
GPT-3.5 (FT SYNT) & 646 & 482 & 84 & 256 & 0.695 & \textbf{0.885} & 0.572 & 0.614\\
GPT-3.5 (FT MBIC) & 484 & 203 & 246 & 535 & 0.683 & 0.663 & 0.704 & 0.694\\
GPT-3.5 (FT MEGA) & 629 & 319 & 101 & 419 & 0.750 & 0.861 & 0.663 & 0.713\\
GPT-3.5 & 384 & 205 & 346 & 533 & 0.582 & 0.526 & 0.651 & 0.624 \\
GPT-4.0 & 393 & 69 & 337 & 669 & 0.659 & 0.538 & \textbf{0.850} & 0.723 \\
Baseline (Random) & 362 & 374 & 368 & 364 & 0.494 & 0.496 & 0.492 & 0.495 \\

\bottomrule
\end{tabular}
\vspace{-10pt}
\end{table}
Table \ref{table:classification_results_BASIL} shows the evaluation results on the BASIL dataset for the same models. Overall, the fine-tuned models all outperformed the non fine-tuned GPT-3.5 on F1-score and recall, while having a worse accuracy and, with one exception (FT MEGA), precision. The fine-tuned models with the more fine-grained bias-type categorizations scored slightly better on average compared to the other fine-tune (FT MBIC) regarding accuracy (65\% vs 62\%), while scoring almost identically on F1-score (39\% vs 40\%) and precision (30\% vs 29\%). For recall, FT MBIC scored higher than the average of the other fine-tunes (59\% vs 63\%), which is in fact the highest score of all evaluated models on this dataset. Overall, GPT-4.0 appears to be the best model, achieving the highest F1-score (44\%), precision (43\%) and accuracy (77\%). One thing clearly noticeable is the high number of false positives on this dataset, leading to rather low scores for f1 and precision, in line with the original BASIL paper \cite{fan-etal-2019-plain}, where the authors fine-tuned two BERT models for the detection of sentences exhibiting "lexical bias" (what they define as "bias stemming from content realization, or how things are said") and "informational
bias" (defined as "sentences or clauses that convey information tangential, speculative, or as background to
the main event in order to sway readers’ opinions
towards entities in the news."). Their lexical bias BERT achieved achieved a precision of 29\%, a recall of 39 \% and a F1-score for 31\%, while the BERT fine-tuned tasked to identify informational BIAS scored 44\% on precision, 43\% on recall and also 43\% on f1. While these results are not directly comparable to ours, due to differences in training and evaluation (e.g the use of two different classifiers for two different categories, each trained on 6819 and evaluated on 400 sentences), their relative similarity to our results might indicate some general tendencies regarding BASIL. The subpar results for bias detection on this dataset may have several reasons, including differences in annotation practices. 
To gain further insights into the effect of the more fine-grained bias-type definitions on result quality, (comparing FT MBIC with the other fine-tunes alone is not enough as FT MBIC uses a different prompt, was fine-tuned on less articles and is based on gpt-3.5-turbo-0613 rather than on gpt-3.5-turbo-1106), another round of evaluation was conducted for the non fine-tuned GPT-3.5. This evaluation was identical to the previous one for GPT-3.5, the only difference was that the fine-grained bias-types definitions were swapped with the coarser ones. On BABE, this resulted in an accuracy of 58\%, a precision of 60\%, a recall of 47\% and a F1-score of 53\%, trailing the GPT-3.5 with the fine-grained type definitions on all four metrics. On BASIL, the results were 70\% for accuracy,  31\% for precision, 40\% for recall and 35\% for F1-score, outperforming the other GPT-3.5 on F1-score and recall while scoring lower for precision and accuracy.
% {"tps": 343, "fps": 224, "fns": 387, "tns": 514, "accuracy": 0.5837874659400545, "precision": 0.6049382716049383, "recall": 0.46986301369863015, "F1-score": 0.5289128758673862}
% {"tps": 328, "fps": 744, "fns": 495, "tns": 2563, "accuracy": 0.7, "precision": 0.30597014925373134, "recall": 0.3985419198055893, "F1-score": 0.3461741424802111}
\begin{table}
\vspace{-10pt}
\caption{Evaluation Results on the BASIL dataset for GPT-3.5-turbo-1106 fine-tuned on BABE, BASIL, Synthetic Data and a combination of all three (MEGA), a previous GPT-3.5 fine-tuned on MBIC, GPT-3.5-turbo-1106 with prompt only and GPT-4-turbo-0125. Best results are highlighted in bold.}
\label{table:classification_results_BASIL}
\begin{tabular}{lrrrrrrrr}
\toprule
\textbf{Model} & \textbf{TP} & \textbf{FP} & \textbf{FN} & \textbf{TN} & \textbf{F1-Score} & \textbf{Recall} & \textbf{Precision} & \textbf{Accuracy}\\
\midrule

GPT-3.5 (FT BABE) & 469 & 1120 & 354 & 2187 & 0.389 & 0.570 & 0.295 & 0.643\\
GPT-3.5 (FT BASIL) & 458 & 1043 & 365 & 2264 & 0.394 & 0.557 & 0.305 & 0.659\\
GPT-3.5 (FT SYNT) & 496 & 1277 & 327 & 2030 & 0.382 & 0.602 & 0.280 & 0.612\\
GPT-3.5 (FT MBIC) & 516 & 1276 & 307 & 2031 & 0.395 & \textbf{0.627} & 0.288 & 0.617 \\
GPT-3.5 (FT MEGA) & 501 & 1062 & 322 & 2245 & 0.412 & 0.609 & 0.320 & 0.664\\
GPT-3.5 & 295 & 654 & 528 & 2653 & 0.332 & 0.358 & 0.311 & 0.714 \\
GPT-4.0 & 366 & 489 & 457 & 2818 & \textbf{0.436} & 0.445 & \textbf{0.429} & \textbf{0.771} \\
Baseline (Random) & 445 & 1853 & 378 & 1454 & 0.285 & 0.541 & 0.193 & 0.460 \\

\bottomrule
\end{tabular}
\vspace{-10pt}
\end{table}
Table \ref{table:classification_results_type_numbers} shows the distribution of identified bias-types by the different models across both datasets. For the sake of readability and representation, only types which were identified more than 100 times by at least one model are included. The table further includes the Jensen–Shannon divergence between the distribution of identifies types and the distribution in the dataset used for fine-tuning as well as the
average Jensen–Shannon divergence between the distribution of identifies types and the
other two datasets (bias-types not included in the table were taken into account for the calculation). Overall, the bias-type distribution of the FT BABE results was the closest to its fine-tuning dataset with a JSD of 0.207. FT BABE also had the greatest difference (0.272) from this value to the average JSD between the distribution of identifies types and the other two individual datasets used for fine-tuning (0.479). The largest part of this difference can be attributed to the high JSD with the synthetically generated fine-tuning dataset, which was 0.688 compared to the JSD with the dataset based on BASIL, which was only 0.27.
Overall, the synthetic fine-tune was responsible for the most chaotic results, among the fine-tuned models. This is firstly visible by the deviating results of FT SYNT, compared to those of FT BABE and FT BASIL, with its values often being outliers for a category. Furthermore, the JSD between the FT SYNT fine-tune dataset type distribution and the distribution of the FT SYNT evaluation results was much higher than the same comparison for FT BABE and FT BASIL, with a JSD of 0.520, which is only 0.034 smaller than the JSD with the two non synthetic datasets. So the model fine-tuned on the synthetic dataset did not only produce vastly different categorizations from the two models which were fine-tuned on non synthetic data (which were relatively close to each other), these categorisations are also vastly different to the ones given in the fine-tune dataset. It is also notable, that FT MEGA, despite being a combination of all three other datasets, was not merely the average between them and that the the non fine-tuned GPT-3.5 was the greatest overall outlier among the models. 
\begin{table}
\vspace{-10pt}
\centering
\caption{Number of identified bias-types on BABE and BASIL for GPT-3.5-turbo-1106 fine-tuned on BABE, BASIL, Synthetic Data and a combination of all three (MEGA), GPT-3.5-turbo-1106 with prompt only and GPT-4-turbo-0125. The greatest outlier from the mean of a row is highlighted in bold. Last two rows show the Jensen–Shannon divergence between the distribution of identifies types and the distribution in the dataset used for fine-tuning as well as the average Jensen–Shannon divergence between the distribution of identifies types and the other two datasets. Average JSD is empty for FT MEGA as it is just a combination of the three individual datasets.}
\label{table:classification_results_type_numbers}
%\begin{tabular}{p{2.5cm}rrrrrrr}
\begin{tabular}{lrrrrrrr}
\toprule
\textbf{bias-type} & \textbf{ FT BABE} & \textbf{FT BASIL} & \textbf{FT SYNT} & \textbf{FT MEGA} & \textbf{GPT-3.5 } & \textbf{GPT-4.0} \\
\midrule

Ad Hominem & 85 & 108 & 453 & 59 & \textbf{456} & 26 \\
Ambiguous Attribution & 2 & 38 & 60 & 3 & \textbf{125} & 7 \\
Emotional Sensationalism & 140 & 98 & 245 & 179 & 234 & \textbf{72} \\
Opinionated & 201 & 154 & 139 & \textbf{434} & 118 & 108 \\
Political & 47 & 484 & \textbf{708} & 246 & 5 & 36 \\
Projection & 9 & 108 & \textbf{157} & 56 & 2 & 30 \\
Source Selection & 12 & 84 & \textbf{165} & 44 & 18 & 8 \\
Unsubstantiated Claims & 22 & \textbf{245} & 21 & 43 & 18 & 38 \\
Word Choice & 916 & 519 & 188 & 973 & \textbf{21} & 775 \\
\hline
JSD (Own FT) & 0.207 & 0.304 & 0.520 & 0.315 & - & - \\
Average JSD (Other FT) & 0.479 & 0.475 & 0.554 & - & - & - \\

\bottomrule
\end{tabular}
\vspace{-10pt}
\end{table}
Despite the instruction to stick to the provided list of bias-types, all models came up with some own. When excluding FT MBIC (which used other definitions) and those hallucinated types, which were chosen less than 3 times, the majority should was already covered by our definitions, like Religious Bias (7, Political), Omission Bias (6, Cherry Picking), False Analogy Bias (6, Faulty Analogy), Appeal to Authority Bias (4, External Validation) and Loaded Language Bias (3, Word Choice). In other cases, the models also came up with completely new types, like Irrelevant Information Bias (3) and Conspiracy Bias (3).
Fine-tuning did not result in a decrease of type hallucinations, despite ensuring that the used datasets did not contain any. FT BABE named an own type in 11 instances, FT BASIL in 21 cases and FT SYNT did so 6 times (Interestingly enough, FT MEGA hallucinated an own type in 11 cases, which is the average of the three other FT models). For comparison, GPT 3.5 chose a new type in 3 cases and GPT-4.0 did so 9 times. 
To gain more insights on type detection, a random sample of BABE and BASIL was manually enhanced by us trough adding the bias-type and the bias strength. Sentences marked as non-biased were not modified in any way. This resulted in 133 biased sentences (21 from BASIL and 122 from BABE). The biased sentences were split across Word Choice (53), Political (25), Opinionated (10), Unsubstantiated Claims (8), Ambiguous Attribution (4), Cherry Picking (4), Emotional Sensationalism (4), Insinuative Questioning (4), Discriminatory (3), Projection (3), Whataboutism (3), Generalization (2), Intergroup (2), Anecdotal Evidence (1), Causal Misunderstanding (1), False Dichotomy (1), Faulty Analogy (1), Mud Praise (1 times), Source Selection (1), Speculation (1) and Straw Man Bias (1). 
Assuming the type distributions of this random sample are somewhat representative for the complete datasets, one might predict that FT BABE, FT MEGA and GPT-4.0 should reach the highest accuracy regarding type classification, as their distributions Section \ref{table:classification_results_type_numbers}, (e.g. highest amount is respectively Word Choice Bias) are the most similar. This hypothesis is confirmed by the actual evaluation, presented in Section \ref{table:Accuracy and Difference}.
\begin{table}
\vspace{-10pt}
\caption{Type detection accuracy and average difference in assigned scores on the manually annotated dataset for GPT-3.5-turbo-1106 fine-tuned on BABE, BASIL, Synthetic Data and a combination of all three (MEGA), GPT-3.5-turbo-1106 with prompt only and GPT-4-turbo-0125. Best results are highlighted in bold.}
\label{table:Accuracy and Difference}
\begin{tabular}{lrr}
\toprule
\textbf{Model} & \textbf{Accuracy(Types)} & \textbf{ Difference(Strengths)} \\
\midrule

GPT-3.5 (FT BABE) & 0.377 & 0.244 \\
GPT-3.5 (FT BASIL) & 0.288 & \textbf{0.223} \\
GPT-3.5 (FT SYNT) & 0.223 & 0.249 \\
GPT-3.5 (FT MEGA) & 0.410 & 0.239 \\
GPT-3.5 & 0.07 & 0.239 \\
GPT-4.0 & \textbf{0.453} & 0.257 \\
Baseline (Random) & 0.045 & 0.308 \\
\bottomrule
\end{tabular}
\vspace{-10pt}
\end{table}
Looking at the output of all models (with the exception of GPT-3.5 as its results were too close to random) combined and excluding types which were checked less than 12 times, the individual types with the highest accuracy were Word Choice Bias (61\%), Emotional Sensationalism Bias (50\%) and Discriminatory Bias (36\%). The types with the lowest accuracy were Ambiguous Attribution Bias (0\%), Projection Bias (0\%) and Insinuative Questioning Bias (6\%). Accordingly, the bias-types (among those rightfully detected at least 5 times) with the lowest average difference to the score assigned in the dataset were Emotional Sensationalism Bias (0.089), Political Bias (0.170) and Discriminatory Bias (0.260). Those with the highest average difference were Opinionated Bias (0.355), Unsubstantiated Claims (0.350) and Word Choice Bias (0.261).
Fine-tuned models perform better with longer articles on the BABE dataset and worse on the BASIL dataset, with consistent relative differences compared to non-fine-tuned models across article lengths.
%Figure \ref{fig:F1-length} displays the achieved F1-scores on both datasets relative to the length of the evaluated articles for all models which were fine-tuned on articles with a length between 10 and 30 sentences (FT-BABE, FT-BASIL, FT-SYNT, FT-MEGA) and the models which were just prompted (GPT-3.5, GPT-4.0). The general trend for BABE shows an increase in F1-scores with longer articles while F1-scores decreased with article length on BASIL. For both datasets, the relative differences between the fine-tuned and the non fine-tuned models F1-scores per article length were overall pretty constant.

\begin{figure}[htbp]
    \vspace{-15pt}
    \centering
    \hspace*{-0.05\textwidth} % Adjust this value as needed
    \includegraphics[width=1.0\textwidth]{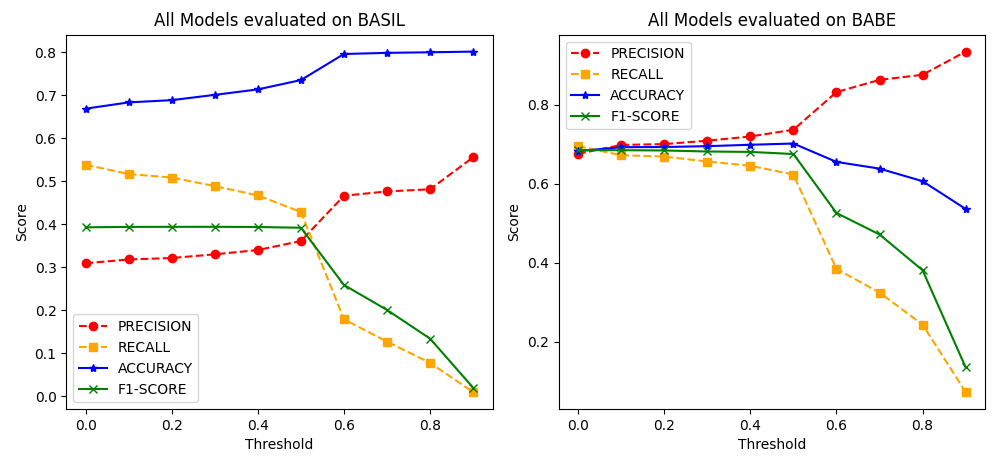} % Adjust this value as needed
    \vspace{-15pt}
    \caption{Evaluation Metrics and Bias Strength Filtering}
    \vspace*{-15pt} % Adjust this value as needed
    \label{fig:f2-filter}
\end{figure}

%\begin{figure}[htbp]
%vspace{-15pt}
 %   \centering
  %  \hspace*{-0.05\textwidth}
   % \includegraphics[width=0.8\textwidth]{related-work/F1_And_Length_Filter_50_cropped.png}
    %\vspace{-10pt}
    %\caption{F1-Score and Article Length }
    %\vspace{-15pt}
    %\label{fig:F1-length}
%\end{figure}
%Looking at \ref{table:classification_results_BABE} and especially \ref{table:classification_results_BASIL}, it appears that false positives are much more common than false negatives.
In order to evaluate if it was possible to decrease the number of false positives based on bias strength scores, the average assigned bias strength was calculated for all true and false positives.
All models on both datasets had average higher scores for their right positives than for their false positives. The largest such difference for BABE was achieved by FT SYNT (0.240), the smallest by FT BABE GPT-3.5 (0.074), with the mean being 0.115.
On BASIL, it was also FT SYNT (0.142) with FT BABE having the smallest difference(0.031). The average here was 0.072. The larger average difference of BABE compared to BASIL could partly explain why there are so many false positives in the related evaluation. %The sentences falsely marked positives are closer to the actual positives on BASIL than on BABE. 
Furthermore, FT SYNT having the highest difference on both datasets could indicate that using (synthetic) data with an even spread of strengths for fine-tuning leads to more realistic relative strength assignments (not to be confused with absolute Table \ref{table:Accuracy and Difference}).
To see if this could be used to improve performance, evaluation results were filtered to change the models decision to "unbiased" for all sentences that were initially marked as "biased" but were assigned a score below a certain threshold. %For readability %(and as every single model had higher strengths for right positives than for their false positives),
%all model classifications were aggregated for both datasets. 
The change for different metrics after applying this filter to the aggregated model classifications is shown in \ref{fig:f2-filter}.
The general trend was the same for both datasets. By filtering out those results with a low bias strength, precision was increased at the expense of recall. This further led to an initial decrease in F1-score and an initially increasing accuracy. Until the point where every sentence with a strength less than 0.5 was filtered out, the increase in accuracy was greater than the decrease in F1-score. Filtering can therefore be seen as a good strategy until this point. However after it, the F1-score started to decrease dramatically while the accuracy either only increased slightly before leveling out or also started to drop. 
Using a majority decision approach with all models on the BABE dataset yielded a higher F1-score (0.736) compared to the average individual scores (0.681), though still lower than the top-performing individual models, while for the BASIL dataset, the majority F1-score (0.405) was also higher than the average (0.391) but remained lower than the best individual models' scores.
%Another possibility to increase performance might be using the majority decision of several models, marking the sentences as biased that were identified by a majority of models in a combination.
%Using all models evaluated on BABE, the achieved F1-score by majority decision was in fact higher (0.736) than the average of their scores (0.681), but still lower than the two of the individual models, with the recall climbing to 0.922, which is higher than every individual score. Precision, on the other hand dropped to 0.613, which was lower than any individual value. Accuracy was slightly lower than the average, with a score of 0.671 compared to 0.682.  
%For Basil, the majority F1-score (0.405) was also higher than the average (0.391), which is again worse than the individual F1-score of two models. Precision was again lower than any individual value (0.270), as was accuracy (0.53). Recall on the other hand was once more higher than any individual value (0.801).
During the writing of this paper, Anthropic released their Claude 3 family, claiming better performance than GPT-4.0 on several benchmarks. While a detailed evaluation was out of scope for this paper, we tested their most powerful Opus model on a subset of BABE containing 148 unbiased and 163 biased sentences as well as a subset of BASIL with 191 unbiased and 50 biased sentences, to get a crude impression of its performance. On BABE, it achieved an accuracy of  71\%, a precision of 85\%, a recall of 55\% and a F1-score of 66\%. On BASIL, it was an accuracy of 78\%, a precision of 47\%, a recall of 36\% and a F1-score of 41\%. While these number would not point to improved capabilities over GPT-4 in terms of bias detection, anecdotal evidence suggests its  textual explanations are better worded. 

\section{Limitations and Ethical Considerations}
\label{sec:limitations}

A generative model developed with the intention
of detecting news media bias-types remains a
generative model: in theory, an adversary could
abuse our model to synthesize biased textual
material. It should also be noted that bias detection is always, to an extent, a subjective matter. What is considered biased and what is not differs depending on who is asked (see the differences between both datasets), therefore no classification will probably ever satisfy everyone at once. Both datasets further had a bias regarding the discussed topics, in that they were very much centered around the US American political discourse. Considering that all used
datasets were published early enough to be incorporated into the training data of
all models, it remains a possibility that certain parts of them were included, even
though none of the models were familiar with datasets with the respective name when queried.  

%limitation evaluation different split of sentences

%%%%%%%%%%%%%%%%%%%%%%%%%%%%%%%%%%%%%%%%%%%%%%%%%%%%%%%%%%%%%%%%%%%%%
% Summary, Conclusions & Future Work
%%%%%%%%%%%%%%%%%%%%%%%%%%%%%%%%%%%%%%%%%%%%%%%%%%%%%%%%%%%%%%%%%%%%%

\section{Summary, Conclusions and Future Work} \label{sec:conclusion}

In this paper, we proposed a fine-grained taxonomy of 27 news bias-types and conducted detailed evaluations on the ability of LLMs to detect bias in news articles. Our experiments gave several insights and provided answers to our research questions:

%\sloppypar{}
\textbf{RQ-1:} Our evaluation generally implies a better performance of the models using the more fine-grained categorization than the one used a coarser one. This is true for the fine-tuned ones as well as when using prompt only.\\[-5mm]

%\sloppypar{}
\textbf{RQ-2:} Contrary to intuition, fine-tuning models with examples that contain the defined bias-types did not lead to fewer hallucinations but more, compared to the models that were not fine-tuned.\\[-5mm]

%\sloppypar{}
\textbf{RQ-3:} While the fine-tuned GPT-3.5 models performed notably better than their non fine-tuned counter part and some fine-tuned models outperformed/were on par with GPT-4.0 on BABE, GPT-4.0 proved to be the most consistent with its relatively good performance, when also accounting for BASIL. It also proved best in identifying the individual bias-types. \\[-5mm]

%\sloppypar{}
\textbf{RQ-4:} Using synthetic data for fine-tuning yielded better results than no fine-tuning at all, while not being too far behind the models fine-tuned on real data. Furthermore, the model fine-tuned on synthetic data differentiated best between right and false positives in terms of assigned bias strength.

Besides answering our four research questions questions, we also presented a novel taxonomy
of 27 news bias-types.
\textit{We are not aware of any sentence-level English news bias detection/categorization model with higher accuracy.}
In future work, we aim to develop models for other
languages.
We also plan to move towards
multi-class classification, as our
experience shows that classes often legitimately overlap.
We plan to experiment with
a hierarchical taxonomy of bias-types.
Finally, we will
migrate to open-source/open-data models.

%\sloppypar{}
\section*{Acknowledgments}
%$\thanks{
\parindent0em
The authors gratefully acknowledge the funding provided by the Free State of Bavaria (``Hitech Agenda Bavaria''). We would also like to thank Michael Reiche for annotation help, the MBICS team for sharing their dataset and three anonymous reviewers for feedback. All views are the authors' own.

% ---- Bibliography ----

%\clearpage

\bibliographystyle{splncs04}
\bibliography{Menzner-Leidner-2024b-NLDB}

\end{document}